\documentclass[10pt, conference]{IEEEtran}
\IEEEoverridecommandlockouts
\usepackage{cite}
\usepackage{amsmath,amssymb,amsfonts}
\usepackage{graphicx}
\usepackage{textcomp}
\usepackage{xcolor}

\usepackage{algorithm}
\usepackage{algpseudocode}
\usepackage{amsmath}
\usepackage{float}
\usepackage{subcaption}
\usepackage{bm}
\usepackage{multirow}
\usepackage{colortbl} 
\usepackage{nicematrix}
\usepackage{xcolor}
\definecolor{myblue}{rgb}{0.2,0.4,0.8}
\newcommand{\Parameters}{\item[\textbf{Hyperparams:}]}
\usepackage[colorlinks=true, linkcolor=myblue]{hyperref}

\def\BibTeX{{\rm B\kern-.05em{\sc i\kern-.025em b}\kern-.08em
    T\kern-.1667em\lower.7ex\hbox{E}\kern-.125emX}}
\usepackage{algorithmicx}
\begin{document}

\title{Hardware-aware Calibrated Clustered Attention for Efficient Visual Geometric Transformers\\
}

\author{
Weitian Wang$^{1,2}$\thanks{Emails: \{weitian.wang, shubham.rai, cecilia.delaparra\}@bosch.com, akash.kumar@ruhr-uni-bochum.de}, Rai Shubham$^1$, Cecilia De La Parra$^1$, Akash Kumar$^2$\\
$^1$Robert Bosch GmbH, Germany,
$^2$Ruhr University Bochum, Germany
}
\maketitle

\begin{abstract}
The Visual Geometry Grounded Transformer (VGGT) marks a significant leap forward in 3D scene reconstruction, as it is the first model that directly infers all key 3D attributes (camera poses, depths, and dense geometry) jointly in one pass. However, this joint inference mechanism requires global attention layers with extremely long sequences that causes a significant latency bottleneck. In this paper, we propose blockwise clustered attention (BC attention) to accelerate the global attention layers in VGGT. By limiting the clustering within HW-friendly neighborhood blocks, BC attention reduces the computation overhead of query clustering as well as the costly data movement between on- and off-chip memory. This enables BC attention to scale to long sequences and deliver practical latency improvements on GPUs. Moreover, we introduce a hashing hyperplane calibration method and a threshold-based error compensation method to reduce clustering errors efficiently, which is a bottleneck in the current clustered attention mechanism. Overall, our experiments on GPU demonstrate that calibrated BC attention accelerates the global attention layers by 2.10-2.63$\times$ and the whole backbone by 1.77-2.35$\times$ with negligible loss (1\%) for large scenes. With a small performance loss (\textless 5\%), calibrated BC attention further achieves a 2.26-2.87$\times$ latency improvement on the global attention layers and a 1.90-2.55$\times$ improvement on the backbone.
\end{abstract}


\section{Introduction}
The Visual Geometry Grounded Transformer (VGGT)~\cite{wang2025vggt} is a recently proposed feed-forward transformer model that directly infers all key 3D attributes of a scene from a variable number of views. By this direct inference, VGGT can outperform state-of-the-art methods while avoiding costly visual-geometry post-processing methods, marking an important breakthrough in 3D computer vision.

\begin{figure}[t]
  \centering
  \includegraphics[width=0.8\linewidth]{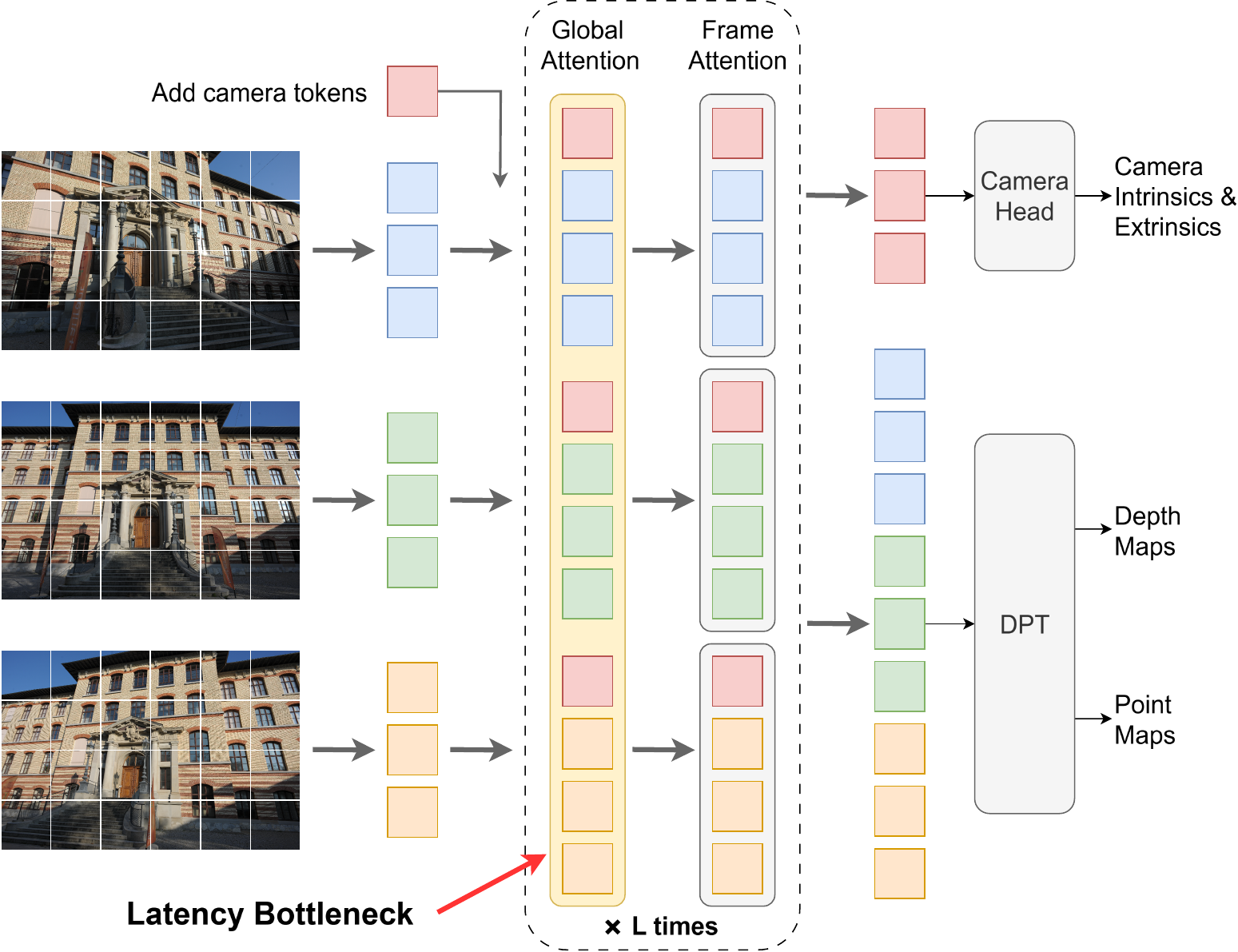}
  \caption{Architecture overview of VGGT~\cite{wang2025vggt}. The global attention layers that perform attention computations across tokens from all views cause a significant latency bottleneck.}
  \label{fig:vggt}
\end{figure}

As shown in Fig~\ref{fig:vggt}, one of the key designs of VGGT is its alternating frame-wise and global attention. In the global attention layers, all tokens from different views participate in the multi-head attention computation. In practice, this approach results in extremely long token sequences (more than 20k tokens) even for small scenes. Thus, the global attention layers become the main latency bottleneck of VGGT, limiting its efficiency on medium and large scene reconstruction.

Motivated by the development of long-context large language models (LLMs) and vision language models (VLMs), many methods~\cite{xu2025xattentionblocksparseattention,zhang2024sparsevlm} have been proposed to ease the high computational costs of long-sequence attention layers. These methods are mainly sparsity-based methods that exploit the pattern that the attention scores of LLMs and VLMs tend to concentrate on a small part of tokens. However, unlike LLMs and VLMs, the attention distributions of VGGT's global attention layers are non-sparse (rare drastic attention scores as shown in Fig.~\ref{fig:attn_dist}), making these sparsity-based methods not ideal.

 On the other hand, this non-drastic attention distribution pattern favors similarity-based methods as the attention weight distributions don't change largely from token to token. Clustered attention\cite{vyas2020fast} was originally proposed to accelerate the attention layers of BERT\cite{devlin2019bert}. Utilizing the similarity between queries, it approximates attention computation by grouping similar queries into clusters and computing attentions only for the cluster centroids; we further refer to this work as \emph{vanilla clustered attention}. Despite its novelty, accelerating modern large vision transformers like VGGT on GPUs with vanilla clustered attention faces several challenges, including large computational and memory access overheads, as well as incompatibility with modern highly efficient attention kernels like FlashAttention~\cite{dao2023flashattention}.

To address these challenges, we propose \emph{blockwise clustered attention} (BC attention) that revisits the idea of clustered attention for VGGT. With hardware-aware algorithmic co-design and optimized GPU implementation, BC attention achieves practical latency improvements on VGGT with negligible performance loss. While our experiments are demonstrated on NVIDIA GPUs, our method is applicable across most types of NVIDIA architectures and can be extended to other accelerator systems. Our main contributions are summarized as follows:
\begin{itemize}
    \item We introduce a hardware-aware blockwise clustering design that lowers the computation overhead and costly on- and off-chip data movement of vanilla clustered attention, enabling practical latency improvements on GPUs for long token sequences.
    \item We propose a novel gradient-based hashing hyperplanes calibration method using differentiable surrogates. Compared with randomly sampled hyperplanes used in vanilla clustered attention, adopting calibrated hyperplanes in the BC attention gives robust results that are more representative of the accurate attention results.
    \item We replace the unstructured top-$k$ error compensation in vanilla clustered attention with a \emph{structured} threshold-based error compensation method that effectively compensates the clustering error. Unlike unstructured attention, these threshold-based structured attention can be efficiently leveraged by high-performance attention kernels such as FlashAttention~\cite{dao2023flashattention}.
\end{itemize}

\section{Related Work}
In this section, we will go through some of the works towards efficiently accelerating attention layers in transformer models. State-of-the-art methods can be broadly classified into: 
\begin{enumerate}
    \item \textbf{{Sparsity-based Attention Acceleration:}}
Sparse attention methods have recently gained popularity as an effective means to address the computational challenges of long-sequence attention. Utilizing the inherently sparse attention mechanism in LLMs, methods like Sparse Transformer~\cite{child2019generating} and BigBird~\cite{zaheer2020big} accelerate inference and extend maximum sequence lengths through local or block-based attention but often require retraining. StreamingLLM~\cite{liu2023ring} enables LLMs to handle unlimited texts without fine-tuning by retaining the attention sinks (initial tokens) and the recent tokens. SparseVLM~\cite{zhang2024sparsevlm} accelerates VLMs through a text-guided training-free token pruning mechanism. For vision transformers (ViTs), SparseViT~\cite{chen2023sparsevit} and MixA-Q~\cite{wang2025mixa} discard or compress unimportant tokens with small magnitudes for latency improvements. However, these methods are all dependent on the sparse attention pattern, which VGGT doesn't exhibit distinctly.

\item \textbf{{Similarity-based Attention Acceleration:}}Another group of methods tries to save computation by clustering or merging similar tokens during attention. ToMe~\cite{bolya2022token} merges similar tokens in vision transformers progressively through a bipartite graph matching algorithm, accelerating ViT without fine-tuning. ToMeSD~\cite{bolya2022token} extends ToMe to accelerate diffusion models~\cite{rombach2022high}. By merging tokens before each component of the block and unmerging them afterwards, ToMeSD reduces compute costs without changing the spatial layout of the tokens. One downside of ToMeSD is that it produces repetitive tokens after merging, potentially limiting the model's expressiveness. On the other hand, clustered attention~\cite{vyas2020fast} clusters similar query tokens for each head using Locality-Sensitive Hashing (SimHash)~\cite{charikar2002similarity} and k-means~\cite{macqueen1967multivariate}. The attention computation is then approximated by using the centroids to represent clustered query tokens. In this way, clustered attention saves computation costs without hurting the uniqueness of tokens. 

\end{enumerate}

\label{sec:related}
In spite of showing promising results on BERT\cite{devlin2019bert}, clustered attention faces challenges when applied to even longer token sequences in VGGT:
\begin{itemize}
    \item  The complexity of clustering $n$ query tokens globally into $k$ clusters with k-means is $O(kn)$. Since the cluster number is proportional to $n$ in practice, the overall clustering overhead grows quadratically with the token length. Additionally, large $n$ and $k$ result in large-sized temporary values like centroids during k-means that can't fit into the SRAM of GPUs, resulting in frequent on- and off-chip data movement.
    \item Vanilla clustered attention recomputes the top-k attention weights for each query, resulting in an unstructured sparse attention pattern that is incompatible with highly optimized GPU kernels such as FlashAttention~\cite{dao2023flashattention}, incurring high memory overhead and reduced efficiency.
\end{itemize}


\section{Calibrated Blockwise Clustered Attention}
\subsection{Observations and Insights}
\begin{figure}[t]
  \centering
  \includegraphics[width=0.7\linewidth]{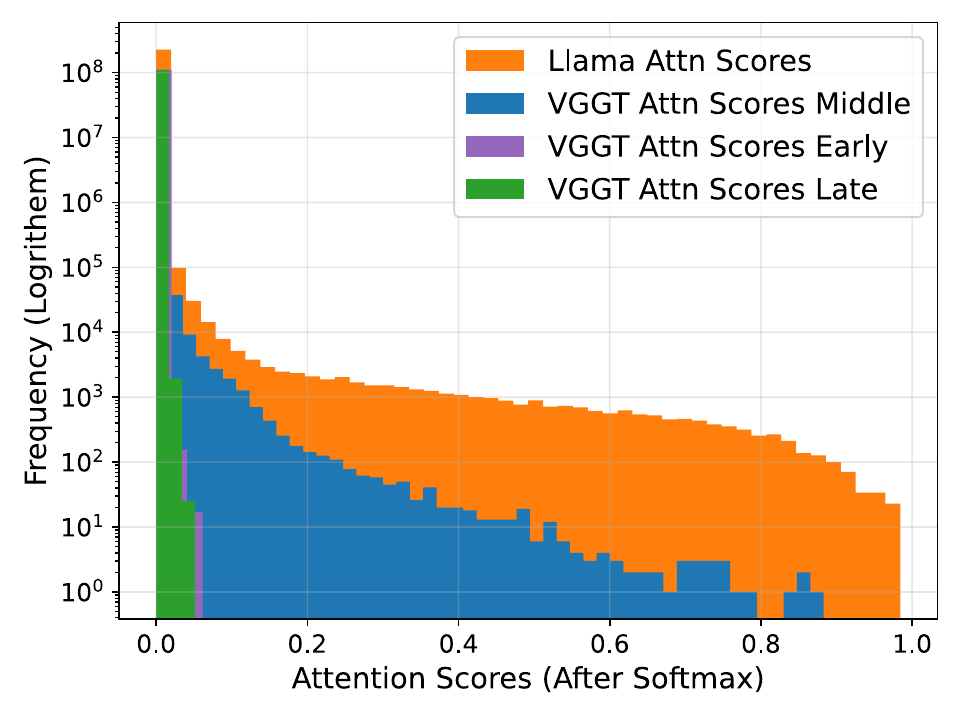}
  \caption{Attention score distribution comparison between VGGT and Llama 3.1 8B\cite{grattafiori2024llama3herdmodels}. VGGT has nearly uniform attention distribution in early and late layers. For middle layers, VGGT still has a more flat attention distribution than Llama.}
  \label{fig:attn_dist}
  \vspace{-2mm}
\end{figure}

Before going into details of BC attention, we want to first present the observations that led us to this solution. In Fig.~\ref{fig:attn_dist}, we show the attention score distributions of global attention layers in VGGT at different depths (3rd,11th,19th) and compare them with the distribution of an LLM (Llama 3.1 8B\cite{grattafiori2024llama3herdmodels}). It can be observed that early and late global attention layers of VGGT have nearly uniform attention score distribution, meaning that each token is attending to all other tokens. For middle layers, the attention starts to concentrate more on certain tokens, but still not as drastically as in Llama, where a vast number of attention scores higher than 0.6 can be observed. From these observations, we infer that sparsity-based attention acceleration methods tailored for LLMs are not ideal for VGGT because there is no small subset of tokens that can accumulate almost all the attention scores, especially for early and late global attention layers. On the other hand, this attention distribution pattern favors similarity-based methods like clustered attention, as the approximation errors are less pronounced for non-dominating tokens.

\begin{figure}[htbp]
  \centering
  \includegraphics[width=0.5\linewidth]{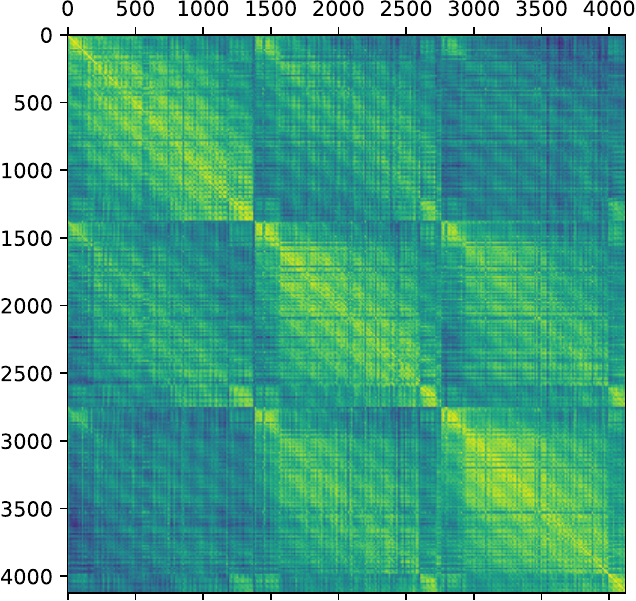}
  \caption{Cosine similarity matrix between queries from the first three frames after adding the positional embedding (RoPE).}
  \label{fig:token_sim}
\end{figure}

Then, we evaluate VGGT models using the cosine similarity metric. Fig. \ref{fig:token_sim} illustrates the cosine similarity matrix between all the tokens in the first three frames of a scene. It can be observed that high similarity scores concentrate near the diagonal, meaning that similar tokens mainly reside in neighborhoods. This pattern arises because, unlike other transformer architectures such as BERT\cite{devlin2019bert} and Stable Diffusion\cite{rombach2022high} which use static positional embeddings added once at the input, VGGT uses Rotary Position Embedding (RoPE) \cite{su2024roformer} applied at every layer. RoPE amplifies positional distinctions, decreasing the similarity between tokens that are far apart. This positional property motivates us to constrain the clustering within blocks to significantly reduce clustering overhead.

\subsection{Hardware-aware Blockwise Clustered Attention}
To understand our proposed approach, it is important to understand vanilla clustered attention. 
Given $\bm{Q},\bm{K},\bm{V} \in \mathbb{R}^{L\times E}$ of a head in an attention layer\cite{vaswani2017attention}, the main purpose of vanilla clustered attention is to reduce the computational complexity of the attention layer by approximating $\bm{X}$ in
\begin{equation}
\bm{X}=\operatorname{softmax}(\frac{\bm{Q}\cdot \bm{K}^T }{\sqrt{d}}) \cdot \bm{V}
\label{eq:attention}
\end{equation}
with $\bm{X^\prime}$ in:
\[
\bm{X^\prime} = \operatorname{scatter}(\operatorname{softmax}(\frac{\bm{Q_c}\cdot \bm{K}^T }{\sqrt{d}}) \cdot \bm{V}, \bm{c_q}) \\
\]
where $\operatorname{scatter}(\bm{M}, \bm{c})_{i,:} \;=\; \bm{M}_{\,c[i],:}$, $\bm{Q_c} \in \mathbb{R}^{k \times E},k<L$ is the $k$ centroids of $L$ query tokens and $\bm{c_q}\in\{0,1,\ldots,k-1\}^L$ is the cluster assignment of $L$ query tokens. The cluster assignments are decided by performing k-means in Hamming space on the hash codes of queries $\bm{H} \in \{0,1\}^L$ obtained using SimHash (More details in Sec.~\ref{sec:hyperplane}). The centroid of each cluster is computed as the mean of all the queries belonging to that particular cluster. In this way, the computational complexity of \eqref{eq:attention} is reduced from $O(L^2E)$ to $O(kLE)$.

Inspired by the observations above and to solve the challenges faced by vanilla clustered attention on long sequences, we propose a hardware-aware blockwise clustered attention (BC Attention) as shown in Alg.~\ref{alg:bc attention}. Instead of the global clustering method used in vanilla clustered attention, BC attention uses blockwise clustering (line 2) as defined in Alg.~\ref{alg:cluster}. By clustering hash codes of queries within blocks of size $B$, blockwise clustering reduces the computational complexity of clustering $L$ tokens into $\lfloor L/\gamma \rfloor$ clusters through $I$ iterations of k-means from $O(IL^2/\gamma)$ to $O(ILB/\gamma)$. Additionally, if we consider GPU implementation, temporary values computed in k-means, like centroids, sample–centroid distance, etc., can fit inside on-chip registers and shared memory. Only one read and one write to HBM (\emph{High Bandwidth Memory}) are required for the clustering of one block. Finally, given that k-means converges faster on smaller problem sizes\cite{davidson2003speeding}, blockwise clustering can converge faster than global clustering, reducing the iteration number $I$ in practice.

\begin{algorithm}[h]
\caption{BC Attention: Approximated Attention Layer with Hardware-aware Block-wise Clustering}
\begin{algorithmic}[1]
\Require $\bm{Q},\bm{K},\bm{V} \in \mathbb{R}^{N \times H \times L \times E}$, block size $B$, number of iterations $I$, compress factor $\gamma$
\Parameters Calibrated hyperplanes $\bm{P_h} \in \mathbb{R}^{H\times 32 \times E}$, layerwise threshold $\tau_\ell$
\Ensure $\bm{X} \in \mathbb{R}^{N \times H \times L \times E}$
    \State $\bm{H} = \Call{SimHash}{\bm{Q}, \bm{P_h}}$
    \State $\bm{C_q} = \Call{BlockCluster}{\bm{H}, B, I, \gamma}$
    \State $k = \lfloor B / \gamma \rfloor$ \Comment{Number of clusters in each block}
    \State $\bm{Q_c}, \bm{M} = \Call{BlockAggregateThresh}{\bm{Q}, \bm{C_q}, B, k, \tau_\ell}$
    \State $\bm{X_c} = \Call{Attention}{\bm{Q_c}, \bm{K}, \bm{V}}$
    \State $\bm{X} = \Call{BlockScatter}{\bm{X_c}, \bm{C_q}, B, k}$
    \State $\bm{X}[\bm{M}] = \Call{Attention}{\bm{Q}[\bm{M}], \bm{K}, \bm{V}}$ \Comment{\small Error compensation}
\end{algorithmic}
\label{alg:bc attention}
\end{algorithm}

\begin{algorithm}[h]
\caption{\textsc{BlockCluster}: Hardware-aware Block-wise Clustering}
\begin{algorithmic}[1]
\Require Hash codes $\bm{H} \in \mathbb{Z}^{N \times H \times L}$, block size $B$, number of iterations $I$, compression factor $\gamma$
\Ensure Cluster assignments $\bm{C_q} \in \mathbb{Z}^{N \times H \times L}$
\State Partition $\bm{H}$ to $N \times H \times \lceil L/B\rceil$ blocks of size at most $B$
\For{each block of hash codes $\bm{h_i}$}
  \State $k = \lfloor B / \gamma \rfloor$ \Comment{Number of clusters in each block}
  \State Load $\bm{H^{(i)}}$ from HBM to on-chip SRAM
  \State Initialize $\bm{h_k}$, $\bm{d}$, $\bm{C_q}^{(i)}$ in SRAM
  \For{$iter=1 \to I$} \Comment{Hamming distance k-means}
    \State Update sample-centroids Hamming distances $\bm{d}$
    \State Update cluster assignments $\bm{C_q}^{(i)}$ according to $\bm{d}$
    \State Update centroids $\bm{h_k}$ by bit-wise majority voting 
    \State Reinitialize centroids of empty clusters if needed
  \EndFor
  \State Write final assignments $\bm{C_q}^{(i)}$ back to $\bm{C_q}$ on HBM
\EndFor
\end{algorithmic}
\label{alg:cluster}
\end{algorithm}

To improve the approximation quality of BC attention (Alg.~\ref{alg:bc attention}), we adopt a headwise calibrated hyperplane during the SimHash (line 1) and recompute the outputs of outliers based on a static layerwise threshold (line 4, 7). Details on these two mechanisms will be demonstrated in the following sections. 


\subsection{Hyperplane Calibration via Differentiable Surrogate}
\label{sec:hyperplane}
\begin{figure}[htbp]
    \centering
    \includegraphics[width=0.95\linewidth]{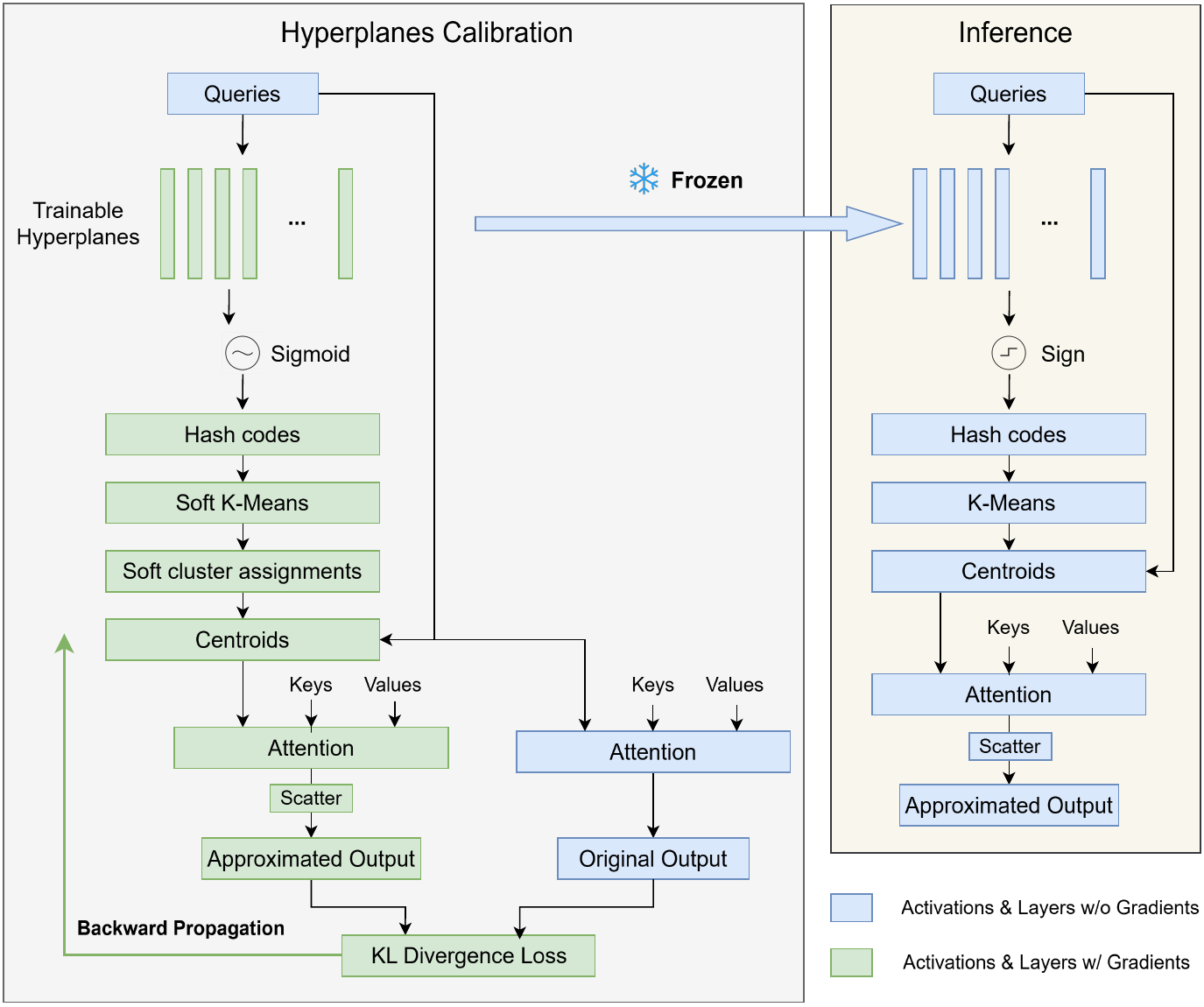}
    \caption{Pipeline of hyperplanes calibration through differentiable surrogate}
    \label{fig:surrogate}
    \vspace{-4mm}
\end{figure}

Similar to vanilla clustered attention, BC attention uses SimHash to generate hash codes that represent the directions of tokens. By representing the $E$ dimension (64 in VGGT) query token
with one \texttt{int32} hash code, SimHash largely reduces the overhead of computing cosine similarities between tokens.

In SimHash, for query tokens $\bm{Q}\in \mathbb{R}^{L \times E}$, A matrix  $\bm{P_h} \in \mathbb{R}^{m \times E}$ that consists of $m$ randomly sampled hyperplanes $\bm{p_i}\ \in \mathbb{R}^E$ is used to project $\bm{Q}$ to generate binary hash codes. For a query token $\bm{q} \in \mathbb{R}^E$, its binary hash vector $\bm{h} \in \{0,1\}^m$ is computed by taking the sign of the dot product between $\bm{q}$ and $m$ hyperplanes, i.e., 
\begin{equation}
\bm{h}= [\operatorname{sign}(\bm{p_0}^T\bm{q}), \operatorname{sign}(\bm{p_1}^T\bm{q}),..., \operatorname{sign}(\bm{p_{m-1}}^T\bm{q})]
\label{eq:hash}
\end{equation}
In practice, we use $m=32$ so that the binary hash vector $\bm{h}$ can be compacted into a single \texttt{int32} hash code, and hash codes of all queries are gathered in $\bm{H} \in \mathbb{Z}^L$. As has been proved in SimHash\cite{charikar2002similarity}, the Hamming distance $d_{ij}$ between $m$-bit hash codes $h_i$,$h_j$ of queries $\bm{q_i}$ and $\bm{q_j}$ has the feature:
\begin{equation}
\operatorname{cos\_sim}(\bm{q_i},\bm{q_j}) = \operatorname{cos}(\pi \cdot \frac{\mathbb{E}[d_{ij}]}{m})
\label{eq:sim}
\end{equation}

This enables clustering query tokens based on similarities efficiently by clustering their hash codes in Hamming distance.

A known issue of SimHash~\cite{charikar2002similarity} is that the randomly sampled hyperplanes lead to results with high variance~\cite{ji2012super}. Our aim is to solve this problem by learning sets of model-dependent hyperplanes on a calibration set so that the hyperplanes can better separate the inputs according to downstream losses without introducing extra overhead during inference. However, neither SimHash nor k-means used in clustered attention are differentiable, and therefore we propose a hyperplane calibration pipeline as described in Fig.~\ref{fig:surrogate}. 

The green pipeline mimics the centroid computation in BC attention with differentiable operations and continues the computation of attention output, which is essentially a gradient estimator of BC attention that \emph{surrogates} it during hyperplane fine-tuning. Hence, we refer to this pipeline as a \textit{differentiable surrogate }. As shown on the left side of Fig.~\ref{fig:surrogate}, by replacing the non-differentiable BC attention layer with the differentiable surrogate, we can optimize the learnable hyperplanes by minimizing the loss between accurate and approximated outputs over a calibration set. After the fine-tuning, we fall back to the original, non-differentiable but efficient BC attention layers as shown on the right side, which now use the calibrated hyperplanes.

Considering the varying activation patterns in different heads and layers, instead of using a single set of 32 randomly sampled hyperplanes, we use a headwise learnable parameter $\bm{P_h} \in \mathbb{R}^{H\times 32\times E}$ for each layer. We replace the random hyperplanes in Eq.~\ref {eq:hash} with our learnable planes in $\bm{P_h}$ and replace the non-differentiable \texttt{sign} function with \texttt{signmoid} (inspired by HashNet~\cite{cao2017hashnet}). Following these modifications, hash codes $\bm{H^{\mathcal{D}}} \in [0,1]^{N\times H\times L\times 32}$ dependent on $\bm{P_h}$ is computed with differentiable operations. Note that $\bm{H^{\mathcal{D}}}$ consists of hash code vectors in 32 dimensions that approximate the 32 bits in hash codes with floating point between 0 and 1.

Subsequently, we adopt k-means with soft assignment~\cite{xie2016unsupervised} to perform differentiable clustering of $\bm{H^{\mathcal{D}}}$. In experiments, we follow the soft assignment implementation from ClusterNet~\cite{chen2019clusternet}.  Soft assignments match samples with clusters via assignment probabilities obtained through a softmax function, which requires a similarity score rather than distances. Following Eq.~\ref{eq:sim}, we define the similarity score of hash code vector $\bm{h_i^{\mathcal{D}}}$,$\bm{h_j^{\mathcal{D}}}$ from $\bm{H}^{\mathcal{D}}$ as:
\[
\operatorname{Sim}_{i,j} = \operatorname{cos}(\pi \cdot \frac{\| \bm{h_i^{\mathcal{D}}}-\bm{h_j^{\mathcal{D}}} \|_1}{m})
\]
For each query block $\bm{Q^{(i)}}$ of size $B$, given a cluster number $k$, the soft k-means calculates a soft assignment $\bm{S}\in [0,1]^{B,k}$ such that $\sum_{j=0}^{k-1}s_{ij}=1$ and $s_{ij}$ represents the assignment probability of token $\bm{q_i}$ belonging to cluster $j$. Hence, we can compute all the centroids of the block $\bm{Q_c^{\mathcal{D}}}\in \mathbb{R}^{k,E}$ differentiably with respect to $\bm{P_h}$ as the weighted average of $\bm{Q^{(i)}}$ according to the assignment probabilities:
\[
    \bm{Q_c^{\mathcal{D}}}=\bm{S}^T\ \cdot \bm{Q^{(i)}}
\]
Afterwards, any loss function $\mathcal{L} = f(\bm{Q_c^{\mathcal{D}}})$ can be used to train the learnable hyperplanes $\bm{P_h}$ through backpropagation. In experiments, we use the KL-divergence between the approximated output of the differentiable surrogate and the real output as the loss to optimize the hyperplanes. Input images from one scene in the test set are used as the calibration inputs.

\subsection{Threshold-based Clustering Error Compensation}
\label{sec:compensation}

As shown in Eq.~\ref{eq:sim}, the Hamming distance of hash codes may fail to accurately reflect similarities due to variance, resulting in poor clustering results for some queries. Hence, error compensation is required for clustered attention methods. Vanilla clustered attention\cite{vyas2020fast} compensates for the error by recalculating the top-$k$ attention scores for each cluster. However, this recomputation involves top-$k$ key index calculation and top-$k$ sparse attention computation for clusters of varying lengths. These unstructured operations are not compatible with attention kernels like FlashAttention~\cite{dao2023flashattention}, which has substantially optimized the structured attention computation on high-performance GPUs. As a result, these operations lead to high latency overheads for long sequences with a large number of clusters.

To efficiently compensate for the outliers during clustering, we propose a threshold-based method that identifies outliers by comparing the Euclidean distance between queries and the centroids of their clusters with a layer-wise threshold. For efficiency, we merge this error compensation with the blockwise centroid aggregation process.


\begin{algorithm}[h]
\caption{\textsc{BlockAggregateThresh}: Blockwise Centroid Aggregation with Outlier Thresholding}
\begin{algorithmic}[1]
\Require $\bm{Q} \in \mathbb{R}^{N \times H \times L \times E}$, cluster assignments $\bm{C_q} \in \mathbb{Z}^{N \times H \times L}$, block size $B$, cluster number $k$, threshold $\tau_\ell$
\Ensure Centroids $\bm{Q_c} \in \mathbb{R}^{N \times H \times K \times E}$, outlier mask $\bm{M} \in \{True,False\}^{N\times H\times L}$
\State Initialize $\bm{Q_c}$ of shape $N,H,\lceil L/B \rceil \cdot k,E$ on HBM
\State Partition $\bm{Q}$ and $\bm{C_q}$ into $N \times H \times \lceil L/B \rceil$ blocks
\For{each block $i$}
  \State Load query block $\bm{Q}^{(i)}$ and its cluster assignments $\bm{C_q}^{(i)}$ from HBM to on-chip SRAM
  \State Initialize centroids $\bm{Q_c}^{(i)}$ of block $i$ in SRAM
  \State Aggregate centroids $\bm{Q_c}^{(i)}$ from $\bm{Q}^{(i)}$ using $\bm{C_q}^{(i)}$
  \For{each query $j$ in $\bm{Q}^{(i)}$}
    \State $c = \bm{C_q}^{(i)}[j]$
    \State $e = \|\bm{Q}^{(i)}[j] -\bm{Q_c}^{(i)}[c] \|_2$ \Comment{Error computation}
    \If{$e$ $> \tau_\ell$} \Comment{Error thresholding}
      \State $\bm{C_q}^{(i)}[j] = -1$
      \State Mark $\bm{Q_c}^{(i)}[c]$ as affected
    \EndIf
  \EndFor
  \State Recompute affected centroids in $\bm{Q_c}^{(i)}$
  \State Write final centroids $\bm{Q_c}^{(i)}$ of block $i$ to $\bm{Q_c}$ on HBM
\EndFor
\State $\bm{M} = \mathbf{1}[\bm{C_q}=-1]$ \Comment{Outlier selection mask of -1}
\end{algorithmic}
\label{alg:agg}
\end{algorithm}

As shown in Alg.~\ref{alg:agg}, we first load query blocks and their cluster assignments from HBM to SRAM to avoid frequent data movements. For each block, cluster centroids are aggregated as the mean of queries in each cluster. The error term is computed in place after the first aggregation and compared with a layerwise threshold. When marked as an outlier, the query is discarded from the centroid computation, and the centroids of clusters with outliers are recomputed. In this way, we merge the error computation and the outlier-free centroids aggregation into computation blocks that can be performed in parallel, and the data movements are avoided as much as possible. This blockwise centroid aggregation returns an outlier mask $\bm{M}$ which is used in Alg.~\ref{alg:bc attention}, line 7, to select outliers for attention computation and fill the corresponding rows in the output. Note that $\bm{M}$ has shape $N\times H\times L$, so different heads can have different numbers of outliers. For computing this multi-head attention with various head lengths efficiently, we use the \texttt{flash\_attn\_varlen\_func} kernel from FlashAttention.

\begin{figure}[htbp]
  \centering
  \includegraphics[width=0.8\linewidth]{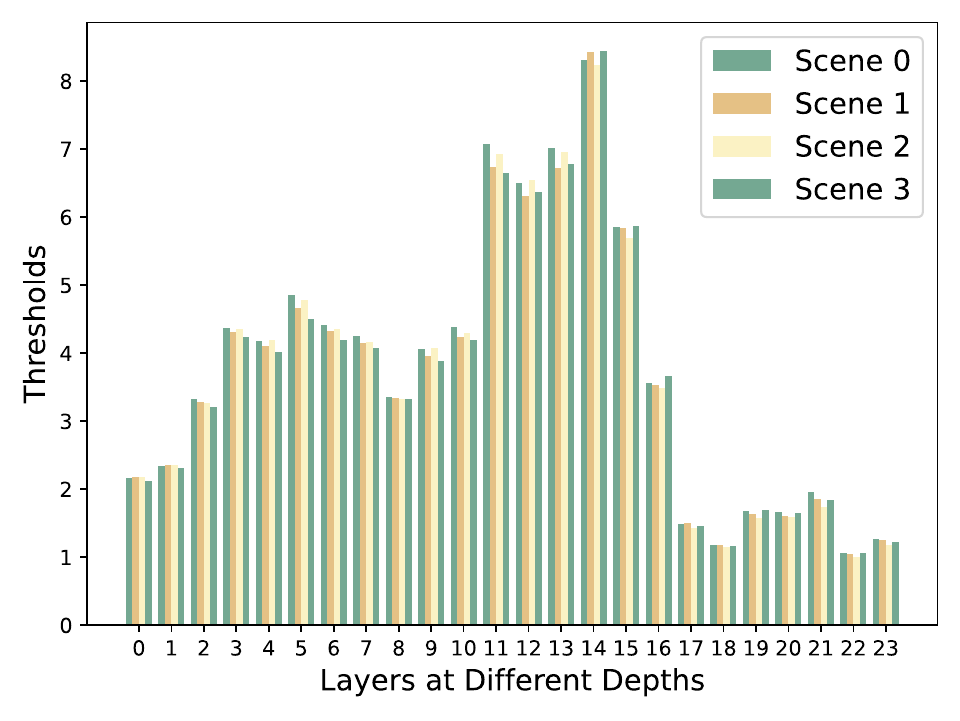}
  \caption{Top 10\% outlier thresholds of 4 random scenes across layers at different depths}
  \label{fig:thresolds}
\end{figure}

To determine the layerwise thresholds, we use a calibration dataset as in Sec.~\ref{sec:hyperplane}. For each layer, the error terms of all queries across all heads are sorted, and the threshold is determined as the 90th percentile. In this way, around 10\% of the queries with the largest error are compensated. Experiments show that the threshold is model-dependent, exhibiting large variation across layers yet consistency across different inputs as shown in Fig.~\ref{fig:thresolds}.

\begin{table*}[hbtp]
\centering
\begin{minipage}{0.55\textwidth}
\centering
\caption{Task Performance on VGGT~\cite{wang2025vggt}}
\begin{tabular}{c|cccccc}
\hline
\rowcolor[HTML]{a3b7ca}
\textbf{Dataset}& \multicolumn{3}{c|}{\textbf{ETH3D}}    & \multicolumn{3}{c}{\textbf{DTU}} \\ \hline
\rowcolor[HTML]{a3b7ca}
\textbf{Metrics}& \textbf{Acc.$\downarrow$} & \textbf{Comp.$\downarrow$} & \textbf{Overall$\downarrow$}& \textbf{Acc.$\downarrow$} & \textbf{Comp.$\downarrow$} & \textbf{Overall$\downarrow$} \\ 
\multicolumn{7}{c}{\cellcolor[HTML]{d1dbe4} Standard Attention} \\ 
Baseline        & 0.871 & 0.529  & 0.700        & 0.448           & 0.437   & 0.443  \\
\multicolumn{7}{c}{\cellcolor[HTML]{d1dbe4} Calibrated BC Attention} \\       
$\gamma = 3$    &\textbf{0.891} &\textbf{0.523} &\textbf{0.707}   &0.528 &0.437 &\textbf{0.481}     \\
$\gamma = 4$    &\textbf{0.930} &\textbf{0.534} &\textbf{0.732}   &0.566 &0.440 &0.503      \\ 
\multicolumn{7}{c}{\cellcolor[HTML]{d1dbe4} Random BC Attention} \\ 
$\gamma = 3$    & 1.080 & 0.563   & 0.821        &0.687   &0.450 & 0.568        \\
$\gamma = 4$    & 1.215 & 0.600   & 0.908        &0.740   &0.453 & 0.597        \\ 
\multicolumn{7}{c}{\cellcolor[HTML]{d1dbe4} ToMeSD with 10\% compensation}      \\ 
$\gamma = 3$    & 1.490 & 0.803   & 1.147        &1.335   &0.479 & 0.907        \\
\hline
\end{tabular}
\label{table:task}
\end{minipage}
\hfill
\begin{minipage}{0.43\textwidth}
\centering

\caption{Latency Comparison}
\begin{tabular}{c|cccl}
\hline
\rowcolor[HTML]{a3b7ca}
\textbf{\#Frames}& \textbf{75} & \textbf{100} & \textbf{150} & \textbf{200} \\
\rowcolor[HTML]{a3b7ca}
\multicolumn{5}{c}{\cellcolor[HTML]{d1dbe4} Backbone with Standard Attention} \\ 
Baseline        & 4.10s   & 6.91s  & 14.56s  & 25.11s         \\ 
\multicolumn{5}{c}{\cellcolor[HTML]{d1dbe4} Backbone with BC Attention} \\ 
$\gamma = 3$    & 2.31s  & 3.50s  & 6.63s   &\multicolumn{1}{r}{\textbf{10.67s (2.35$\times$)}}   \\
$\gamma = 4$    & 2.16s  & 3.28s  & 6.13s   &\multicolumn{1}{r}{\textbf{9.84s (2.55$\times$)}}    \\ 
\multicolumn{5}{c}{\cellcolor[HTML]{d1dbe4} Standard Global Attention Layers} \\ 
Baseline        & 3.40s  & 5.97s  & 13.18s  &23.27s         \\
\multicolumn{5}{c}{\cellcolor[HTML]{d1dbe4} BC Global Attention Layers} \\ 
$\gamma = 3$    & 1.62s  & 2.59s  & 5.26s   &\textbf{8.84s (2.63$\times$)}         \\
$\gamma = 4$    & 1.50s  & 2.39s  & 4.82s   &\textbf{8.11s (2.87$\times$)}       \\ \hline
\end{tabular}

\label{table:latency}
\end{minipage}
\end{table*}

\begin{table}[h]
    \centering

    \caption{Task Performance on MapAnything~\cite{keetha2025mapanything}}
    \begin{tabular}{c|ccc}
    \hline
    \rowcolor[HTML]{a3b7ca}
    \textbf{Dataset}& \multicolumn{3}{c|}{\textbf{ETH3D}} \\ \hline
    \rowcolor[HTML]{a3b7ca}
    \textbf{Metrics}& \textbf{Acc.$\downarrow$} & \textbf{Comp.$\downarrow$} & \textbf{Overall$\downarrow$} \\ 
    \multicolumn{4}{c}{\cellcolor[HTML]{d1dbe4} Standard Attention} \\ 
    Baseline        & 0.114 & 0.125  & 0.120        \\
    \multicolumn{4}{c}{\cellcolor[HTML]{d1dbe4} Calibrated BC Attention} \\       
    $\gamma = 4$    &\textbf{0.115} &\textbf{0.127} &\textbf{0.121}       \\ 
    \hline
    \end{tabular}
    \label{tab:task_ma}
\end{table}

\vspace{-1em}
\section{Experiments}

\begin{figure*}[htbp]
    \centering
    \begin{subfigure}{0.32\textwidth}
        \includegraphics[width=\linewidth]{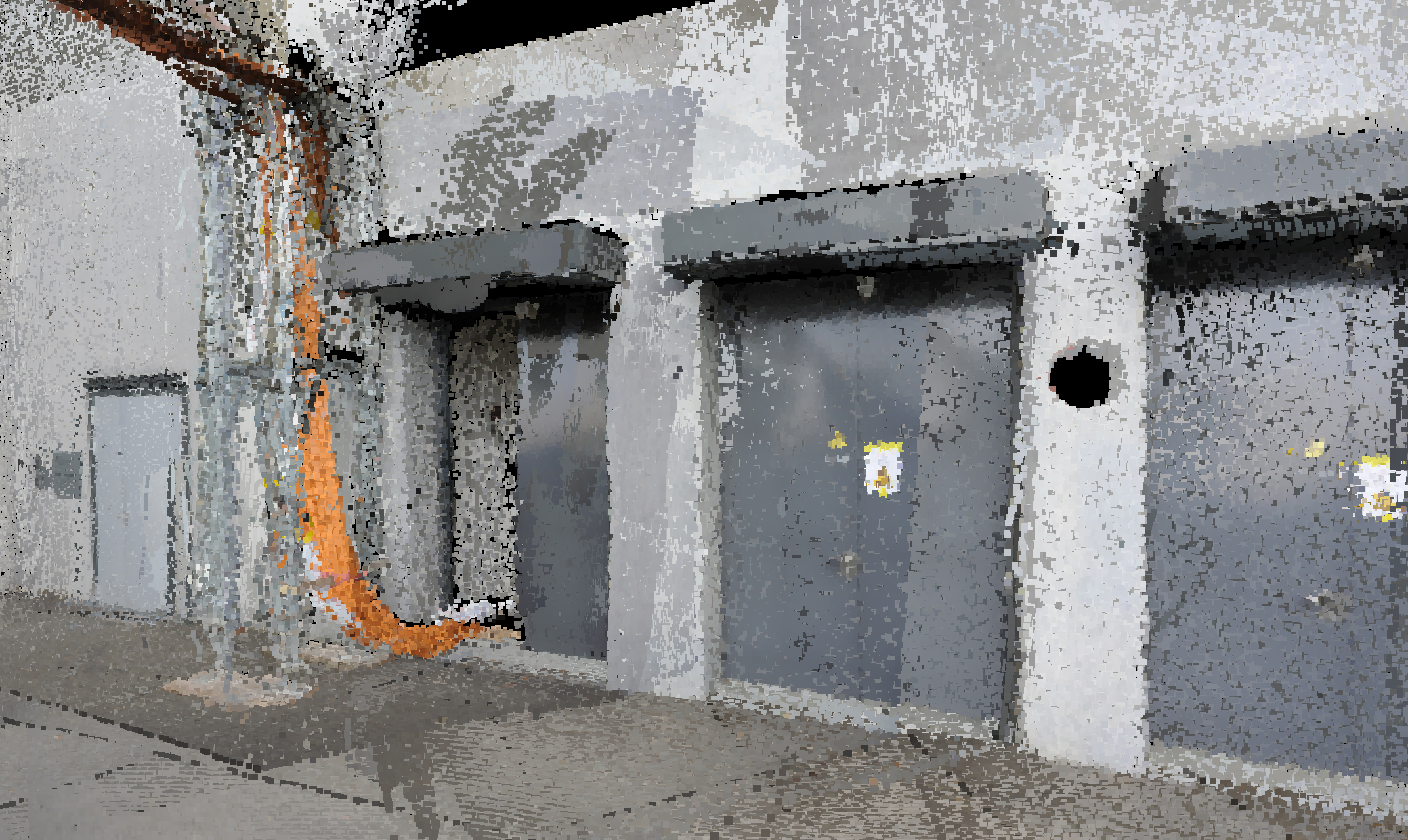}
        \caption{Standard Attention (Baseline)}
    \end{subfigure}
    \begin{subfigure}{0.32\textwidth}
        \includegraphics[width=\linewidth]{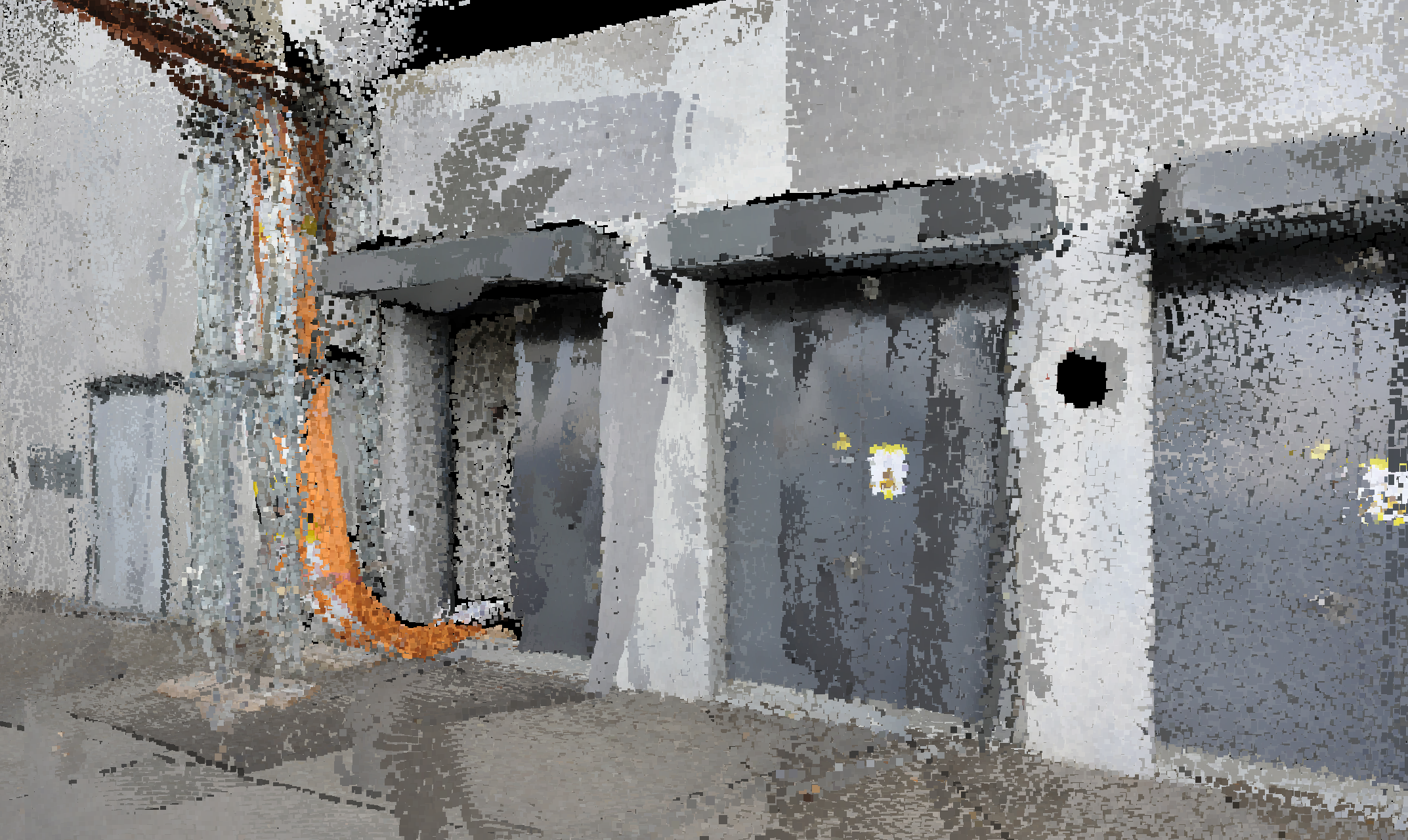}
        \caption{Calibrated Blockwise Clustered Attention}
    \end{subfigure}
    \begin{subfigure}{0.32\textwidth}
        \includegraphics[width=\linewidth]{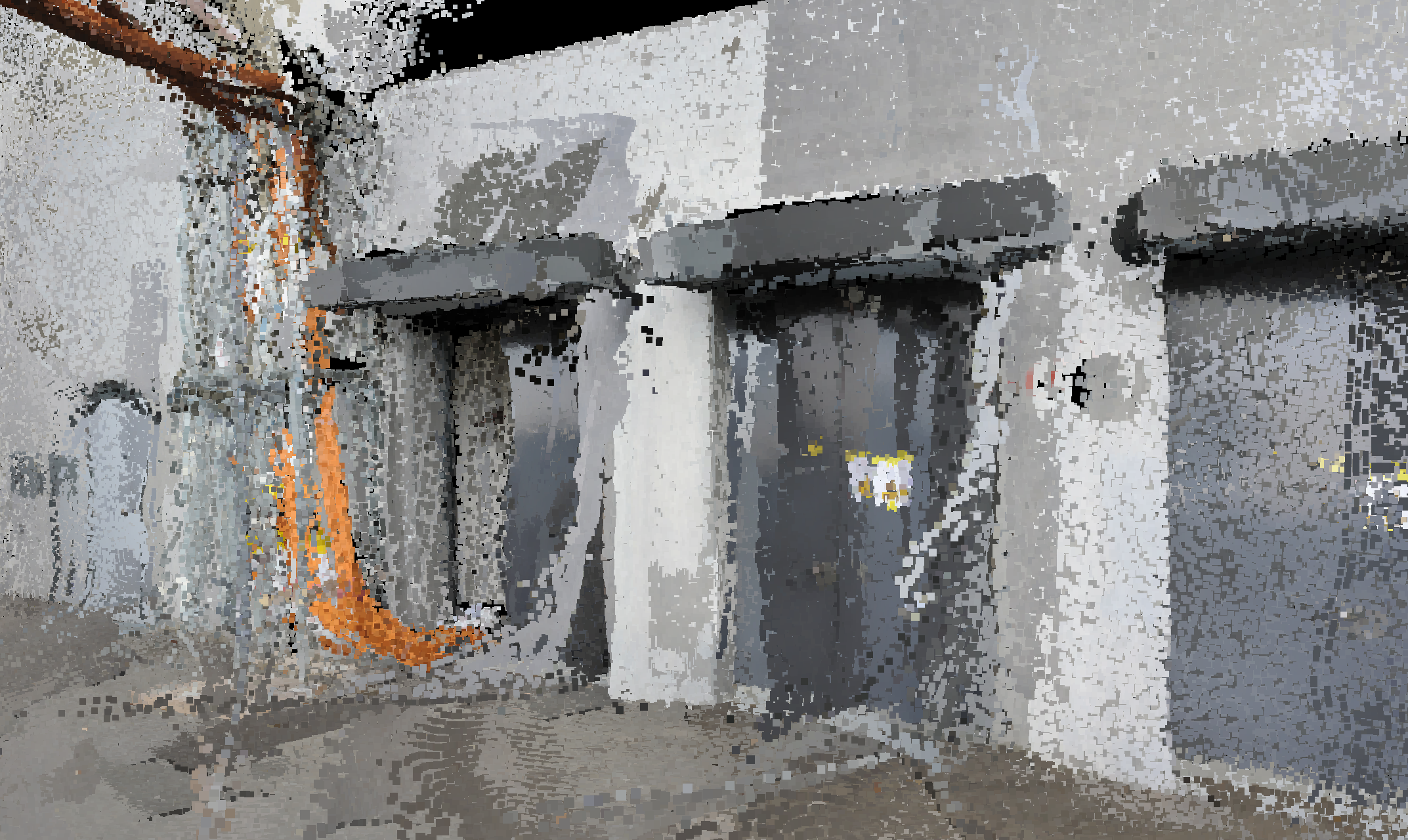}
        \caption{Random Blockwise Clustered Attention}
    \end{subfigure}
    \caption{Comparison of reconstruction results using standard attention, blockwise clustered attention with calibrated and random hyperplanes. The reconstruction result using random hyperplanes exhibits noticeable noise compared to the other two results.}
    \vspace{-1mm}
\label{fig:reconstruct compare}
\end{figure*}

\subsection{Approximation Evaluation}
\label{sec:performance}
We replace the 24 global attention layers in the pretrained VGGT-1B model with our BC attention and compare its task performance on ETH3D~\cite{schoeps2017cvpr} for point map estimation and DTU~\cite{jensen2014large} for dense Multi-View Stereo (MVS) estimation with the standard VGGT. For point map estimation on ETH3D, we use the depth and camera heads (i.e., unprojecting the predicted depth maps to 3D using the predicted camera parameters) because it yields higher accuracy, as stated in the VGGT~\cite{wang2025vggt} paper. We report the Accuracy, Completeness, and Overall (Chamfer distance) for both tasks. Note that these metrics are relative distances that depend on the alignment and thresholding of point clouds, so the numbers we report here are not exactly the same as in the original VGGT\cite{wang2025vggt} paper. However, we employ the same point cloud processing method for all the experiments for fairness.

As shown in Table~\ref{table:task}, on ETH3D, VGGT with calibrated BC attention shows negligible overall performance loss (1\%) with a compression factor $\gamma = 3$ (compress the query length by 3$\times$) and small overall performance loss (\textless 5\%) with a compression factor $\gamma = 4$. On the other hand, when using random hyperplanes, BC attention introduces rather large loss (\textgreater 17\%). We show one example in Fig.~\ref{fig:reconstruct compare} for qualitative evaluation. It can be observed that the reconstruction result using random hyperplanes exhibits noticeable noise compared to the baseline and calibrated results. 

The DTU dataset contains ground-truth geometry of small objects obtained with a structured-light scanner at \emph{millimeter-level} accuracy under controlled lab conditions. We evaluate BC attention on DTU to show that the overall performance loss introduced by calibrated BC attention is negligible (\textless 0.05mm) under this high-fidelity setup as well.

In the last row of Table~\ref{table:task}, we also compared BC attention with ToMeSD~\cite{bolya2023token}. We use ToMeSD to merge similar tokens in each frame before attention computation. For fairness, we also recompute the top 10\% outliers during merging. It can be observed that ToMeSD gives worse results with compression factor $\gamma=3$ than random BC attention with $\gamma=4$.

To show BC attention's versatility in accelerating different models with similar attention design, we evaluate it on the MapAnything~\cite{keetha2025mapanything} model, a follow-up work of VGGT using a similar \emph{alternating frame/global attention design}. Notably, while MapAnything does not apply RoPE in the attention layers, its tokens similarity patterns are still similar to those of VGGT. As shown in Table~\ref {tab:task_ma}, MapAnything with BC attention maintains similar results to the standard one.

\vspace{-0.5em}
\subsection{Latency Performance}
\label{sec:latency}
In this section, we evaluate BC attention's latency improvement on hardware. We use the NVIDIA H200 GPU as the evaluation hardware. All experiments are run with PyTorch~\cite{paszke2019pytorch} in Bfloat16 precision, and FlashAttention2~\cite{dao2023flashattention} is used as the attention kernel for both standard and BC attention.

As shown in Table~\ref{table:latency}, we evaluate the latency improvement of BC attention compared to standard VGGT with 75 to 200 input frames, which is a usual range in 3D reconstruction tasks to guarantee adequate viewpoint coverage and geometric consistency of large scenes. Due to the clustering overhead and the latency of other parts of the model, the acceleration brought by BC attention becomes more evident for larger scenes.

With a compression factor $\gamma=3$, BC attention achieves 2.10$\times$ to 2.63$\times$ latency improvement on all 24 global attention layers during inference, and 1.77$\times$ to 2.35$\times$ improvement on the whole backbone. With $\gamma=4$, BC attention further brings 2.26$\times$ to 2.87$\times$ improvement on the global attention layers and 1.90$\times$ to 2.55$\times$ on the backbone.
\section{Ablation Studies}
\subsection{Error compensation}
To validate the importance of error compensation, we reduce the error compensation rate from 10\% to 5\%. This increases the accuracy distance on ETH3D from 0.891 to 1.320, highlighting the necessity of sufficient error compensation for maintaining the approximation quality of BC attention.

\subsection{Block Sizes Choice}
In experiments in Sec.~\ref{sec:performance} and Sec.~\ref{sec:latency}, we use block size $B=128$. In this section, we investigate the impact of different block sizes on BC attention's approximation quality. 
\begin{table}[h]
\centering
\caption{Task Performance using Different Block Size}
\begin{tabular}{l|ccc}
\hline
Block Size & 64      & 128     & 256 \\ \hline
ETH3D Acc. $\downarrow$ & 0.906   & 0.891   & 0.129    \\ \hline
\end{tabular}
\end{table}

In the above table, we list the reconstruction accuracy on ETH3D using different block sizes with the same number of k-means iterations. It can be observed from the table above that using block sizes of 64 and 128 has similar task performance. However, a bigger block size of 256 results in worse performance. We speculate that a bigger block size needs more iterations to converge during k-means.

\section{Conclusion}
In this work, we proposed hardware-aware blockwise clustered attention (BC attention). By constraining clustering within local HW-friendly blocks, calibrating hash-based hyperplanes through a differentiable surrogate, and introducing threshold-based error compensation, our method achieves both accuracy robustness and practical GPU acceleration. Experiments show that calibrated BC attention delivers 2.1–2.6× speedup on global attention layers and 1.8–2.4× end-to-end backbone acceleration with negligible accuracy loss (1\%). As our proposed approach focuses solely on local token similarities, future work could further explore exploiting framewise similarities, especially in use cases with input frames having temporal relations.

{
    \small
    \bibliographystyle{IEEEtran}
    \bibliography{main}
}

\end{document}